\documentclass[runningheads]{llncs}
\usepackage{float}
\usepackage{graphicx}
\usepackage{amsmath}
\usepackage{amssymb}
\usepackage{hyperref}
\usepackage{booktabs}
\usepackage{multirow}
\begin{document}


\title{Why Registration Quality Matters: Enhancing sCT Synthesis with IMPACT-Based Registration}

\author{Valentin Boussot\inst{1}\orcidID{0009-0003-2465-5458} \and
Cédric Hémon\inst{1}\orcidID{0009-0003-6669-5108} \and
Jean-Claude Nunes\inst{1}\and
Jean-Louis Dillenseger\inst{1}}
\authorrunning{V. Boussot et al.}
\titlerunning{Enhancing sCT Synthesis with IMPACT-Based Registration}
%
\institute{Univ Rennes 1, CLCC Eug\`ene Marquis, INSERM, LTSI - UMR 1099, F-35000 Rennes, France \\
\url{https://ltsi.univ-rennes.fr/}}
\maketitle              
\begin{abstract}
We participated in the SynthRAD2025 challenge (Tasks 1 and 2) with a unified pipeline for synthetic CT (sCT) generation from MRI and CBCT, implemented using the KonfAI framework. Our model is a 2.5D U-Net++ with a ResNet-34 encoder, trained jointly across anatomical regions and fine-tuned per region. The loss function combined pixel-wise L1 loss with IMPACT-Synth, a perceptual loss derived from SAM and TotalSegmentator to enhance structural fidelity. Training was performed using AdamW (initial LR = 0.001, halved every 25k steps) on patch-based, normalized, body-masked inputs ($320 \times 320$ for MRI, $256 \times 256$ for CBCT), with random flipping as the only augmentation. No post-processing was applied. Final predictions leveraged test-time augmentation and five-fold ensembling. The best model was selected based on validation MAE. Two registration strategies were evaluated: (i) Elastix with mutual information, consistent with the challenge pipeline, and (ii) IMPACT, a feature-based similarity metric leveraging pretrained segmentation networks. On the local test sets, IMPACT-based registration achieved more accurate and anatomically consistent alignments than mutual information–based registration, resulting in improved sCT synthesis with lower MAE and more realistic anatomical structures. On the public validation set, however, models trained with Elastix-aligned data achieved higher scores, reflecting a registration bias favoring alignment strategies consistent with the evaluation pipeline. This highlights how registration errors can propagate into supervised learning, influencing both training and evaluation, and potentially inflating performance metrics at the expense of anatomical fidelity. By promoting anatomically consistent alignment, IMPACT helps mitigate this bias and supports the development of more robust and generalizable sCT synthesis models.

\end{abstract}

\section{Introduction} \label{sec:intro}

The generation of synthetic computed tomography (sCT) images from magnetic resonance imaging (MRI) or cone-beam CT (CBCT) supports a wide range of clinical workflows, including MRI-only radiotherapy planning and longitudinal follow-up using low-dose imaging. sCT aims to produce CT-like images with accurate Hounsfield units, closely resembling diagnostic-quality CT, thereby enabling dose calculation, image registration, anatomical segmentation, and treatment planning.

When derived from MRI, sCT eliminates the need for an additional CT scan, thus avoiding exposure to ionizing radiation, a particularly important benefit for pediatric patients and those requiring repeated imaging. When derived from CBCT, sCT is used to overcome the modality’s technical limitations (such as artifacts, noise, and inaccurate Hounsfield units), which are inherent to its low-dose acquisition protocol.

However, producing high-quality sCT images remains challenging due to fundamental differences in acquisition physics, intensity distributions, and anatomical visibility across imaging modalities. This task is further complicated by variability in scanner hardware, acquisition sequences, and imaging protocols, which introduce substantial heterogeneity in the training data. Such variability results in inconsistencies in contrast, resolution, and anatomical representation within the same modality \cite{spadea2021deep}. Deep learning methods, particularly supervised encoder-decoder architectures, have recently shown strong potential for cross-modal image synthesis by learning direct mappings between MRI or CBCT and CT \cite{spadea2021deep,zhong_clinical_2024,sherwani_systematic_2024}. When high-quality paired data are available, supervised methods consistently outperform unsupervised approaches, offering superior anatomical accuracy and intensity realism in the synthesized CT images \cite{rossi2021comparison}.

We followed the prevailing trend by adopting a supervised learning approach for synthetic CT generation, leveraging an encoder–decoder architecture that has shown strong performance in medical image synthesis tasks. A fundamental requirement for such supervised methods is the availability of spatially aligned image pairs. However, in clinical practice, multimodal images are rarely perfectly aligned due to inter-session variations in patient positioning and anatomy \cite{florkow2019impact,yang2020unsupervised}, particularly in abdominal and thoracic regions, which are highly sensitive to breathing, organ motion, and filling effects. Consequently, voxel-level misalignments are common and introduce anatomical inconsistencies into the training data. These alignment errors are especially critical in supervised learning, as it can significantly degrade model performance and lead to anatomically implausible or unrealistic predictions. Achieving accurate multimodal registration remains challenging, especially for MRI, where geometric distortions, variable soft-tissue contrast, and the absence of bone signal limit the effectiveness of conventional intensity-based algorithms such as Elastix \cite{klein2009elastix}.

To overcome these challenges, feature-based registration methods such as MIND \cite{heinrich2012mind} and IMPACT \cite{boussot2025impact} have been proposed. These approaches compare spatially structured representations rather than raw intensities, enabling more robust and anatomically consistent alignment across modalities. Unlike handcrafted descriptors such as MIND, IMPACT leverages deep semantic features extracted from pretrained segmentation models. By exploiting the spatial Jacobian of the feature maps, IMPACT introduces a differentiable loss that encourages anatomical correspondence even under severe appearance shifts, making it particularly effective in this context. 

In this study, we propose a unified deep learning pipeline for synthetic CT generation from MRI and CBCT, developed and evaluated in the context of the SynthRAD2025 challenge \cite{thummerer2025synthrad2025}. Our main contributions are as follows:
\begin{itemize}
\item We investigate the impact of registration quality on supervised image synthesis, comparing traditional Elastix-based alignment with a semantic loss-guided approach leveraging the IMPACT metric.
\item We introduce IMPACT-Synth, a novel perceptual loss based on the Segment Anything Model (SAM) \cite{kirillov2023segment}.
\end{itemize}

Our results highlight that accurate intermodal registration of the training data is critical for generating structurally faithful synthetic CTs. Misaligned training pairs introduce systematic spatial biases that the model may exploit to artificially boost image similarity metrics. However, this gain comes at the expense of anatomical plausibility, often leading to structurally inconsistent or unrealistic outputs. These findings underscore the importance of proper registration as a prerequisite for reliable cross-modal image synthesis.

\section{Method} 
We participate in both tasks of the SynthRAD2025 challenge: MRI to CT and CBCT to CT synthetic image generation, using a unified pipeline based on supervised training of an encoder–decoder architecture optimized with a combination of L1 loss and perceptual losses.

\section{Data} 

\subsection{Pre-alignment process}  

Two registration approaches were investigated:

\begin{itemize}
  
\item \textbf{Baseline registration}: This approach used the Elastix parameter files provided in the official SynthRAD2025 GitHub repository. To reduce the impact of large anatomical mismatches between CT and the secondary modality (MRI or CBCT), a manual curation step was applied. A subset of misaligned image pairs was excluded from the training set, following the same strategy as the one used for test set construction in the challenge. As a result, 101 patients were removed from Task 1 and 50 from Task 2.
\item \textbf{IMPACT-based registration} \cite{boussot2025impact}: This approach relies on a custom setup using the IMPACT semantic similarity metric. For MRI-to-CT registration, the TotalSegmentator-based model \texttt{M730} was used with features extracted from the 7\textsuperscript{th} layer in static mode, combined with the MIND loss. For CBCT-to-CT registration, the same model was applied in Jacobian mode, using features from the 2\textsuperscript{nd} layer. All registrations were performed using a three-level resolution pyramid and a final B-spline grid spacing of 10~mm. The resulting B-spline transformations are publicly available at: \url{https://huggingface.co/datasets/VBoussot/synthrad2025-impact-registration}.
\end{itemize}

\subsection{Preprocessing.} 
After registration, all images were processed within the patient body mask, with voxels outside the mask set to $-1$. Preprocessing was modality-specific:
\begin{itemize}
    \item \textbf{CT:} intensities were clipped to a fixed range \([-1024, 3071]\) and linearly normalized to the range \([-1, 1]\).
    \item \textbf{CBCT:} intensities were clipped between the minimum and the 99.5th percentile inside the mask, then normalized to \([-1, 1]\).
    \item \textbf{MRI:} intensities were standardized within the mask using zero mean and unit variance, without any intensity clipping.
\end{itemize}

\subsection{Training and Validation Strategy}

For both tasks, we adopted the same methodology based on a nested five-fold cross-validation strategy during the development phase. To simulate a local test set, a subset of 75 images was initially held out. The remaining data were used in an inner five-fold cross-validation to tune hyperparameters and select the optimal model configuration. Once the training strategy was finalized, we retrained five models on the full training set, reintegrating the 75 previously held-out images, using a standard five-fold split. The final submission was obtained by ensembling the predictions of these five models.

\section{Model} \label{sec:model}

\subsection{Spatial Configuration}

We adopted a 2.5D strategy for both tasks, in which each input sample is composed of the target slice stacked with a number of adjacent slices to provide contextual information. For CBCT-to-CT synthesis, we used one slice before and one slice after the target, resulting in a three-channel input. For MR-to-CT synthesis, we used two slices before and two slices after, resulting in a five-channel input.

\subsection{Network Architecture}

Our synthesis pipeline is built upon an encoder–decoder architecture, specifically a U-Net++ \cite{zhou2018unet++} with a ResNet-34 encoder \cite{he2016deep}. U-Net++ extends the classical U-Net by introducing nested and dense skip connections, which enhance feature propagation, reduce the semantic gap between encoder and decoder feature maps, and facilitates gradient flow during training.

The encoder is composed of residual blocks from ResNet-34, each consisting of convolutional layers followed by batch normalization and ReLU activations. The decoder includes upsampling layers and dense skip connections from multiple encoder depths, as defined by the U-Net++ architecture. This design enables richer multi-scale feature fusion and supports the learning of more detailed anatomical structures.

The final architecture contains approximately 26.07 million trainable parameters.

\subsection{Implementation Details}

All experiments were implemented using KonfAI \cite{boussot2025konfai}, our in-house deep learning framework built on top of PyTorch v2.6.0+cu124. KonfAI offers a modular and fully configurable training pipeline, driven by YAML-based experiment management. It natively supports 2.5D patch-based processing, model ensembling, and test-time augmentation, enabling reproducible and scalable experimentation across tasks.

\section{Training} \label{sec:training}

\subsection{Training Strategy}

The model was trained on paired input–target images. For each fold, we first trained a global model from scratch using random normal weight initialization with a standard deviation (gain) of 0.02, covering all anatomical regions. Subsequently, region-specific models were fine-tuned from the global model using the checkpoint that achieved the best performance in terms of MAE on the validation set. Fine-tuning was performed jointly for the abdomen and thorax (AB+TH), while the head and neck (HN) region was treated independently.

Patch-based training was employed, with a patch size of $256 \times 256$ for Task~2 and $320 \times 320$ for Task~1, using a batch size of 32. Both the initial training and the fine-tuning phases started with a learning rate of 0.001. A StepLR decay schedule was applied, reducing the learning rate by a factor of 2 every 25,000 steps. Early stopping was applied when the validation performance did not improve for 25,000 consecutive steps. All configurations for Task 1 and Task 2 are publicly available on GitHub: \href{https://github.com/vboussot/Synthrad2025_Task_1}{Task 1} and \href{https://github.com/vboussot/Synthrad2025_Task_2}{Task 2}.

\subsection{Data Augmentation}

Only one random flip per image was applied during training, refreshed at each epoch, which was sufficient to prevent abrupt overfitting. This minimal augmentation strategy also enabled TTA using flipping at inference time.

\subsection{Model Selection}

For each configuration, two checkpoints were retained: the model achieving the best performance on the validation set, and the final epoch model. Empirically, the final model often yielded superior results on the held-out test set.

\subsection{Inference and Ensembling}

To enhance prediction robustness and generalization, we employed test-time augmentation (TTA) based on image flipping. Predictions were computed from flipped inputs, then inverted back to the original orientation before being averaged to produce the final output. This was combined with a cross-validation ensemble, where predictions from models trained on different folds were averaged to produce the final output. This ensembling strategy reduces variance and mitigates overfitting to individual data splits, leading to more stable and consistent results.

\subsection{Loss Functions}

All models were trained using the MAE loss as a baseline. In addition, we explored perceptual loss functions inspired by the IMPACT registration framework \cite{boussot2025impact}. These losses aim to incorporate semantic consistency by leveraging features extracted from pretrained segmentation networks. Specifically, we evaluated three types of perceptual losses:
\begin{itemize}
    \item The original VGG-based perceptual loss, computed from intermediate features of a VGG-19 network pretrained on ImageNet.
    \item A perceptual loss based on features extracted from the SAM \cite{kirillov2023segment} (Segment Anything Model) encoder.
\end{itemize}

\section{Evaluation} \label{sec:evaluation}

\subsection{Evaluation Metrics}

To evaluate the quality of sCT generation in the SynthRAD2025 challenge, we computed the full suite of image similarity metrics defined by the challenge organizers. These include:

\begin{itemize}
  \item \textbf{Mean Absolute Error (MAE)}: average absolute Hounsfield unit difference between sCT and CT.
  \item \textbf{Peak Signal-to-Noise Ratio (PSNR)}: measures the ratio between the maximum possible image intensity and the MSE error.
  \item \textbf{Multi‑Scale Structural Similarity Index (MS‑SSIM)}: an advanced version of SSIM computed at multiple image scales, reflecting perceptual image quality.
\end{itemize}

\subsection{Model Selection Strategy}

We defined the best model based on comprehensive performance across all evaluation metrics (MAE, PSNR, MS‑SSIM). Among candidate checkpoints, the model exhibiting the highest aggregate image similarity was selected as the final submission.

\section{Results} \label{sec:results}

\begin{table}[H]
\centering
\caption{Comparison of synthesis performance between Baseline and IMPACT registration on Task 1 and Task 2 of the local test set.}
\begin{tabular}{llcccc}
\toprule
\textbf{Region} & \textbf{Metric} & \multicolumn{2}{c}{\textbf{Task 1}} & \multicolumn{2}{c}{\textbf{Task 2}} \\
\cmidrule(lr){3-4} \cmidrule(lr){5-6}
& & \textbf{Baseline} & \textbf{IMPACT} & \textbf{Baseline} & \textbf{IMPACT} \\
\midrule
\multirow{3}{*}{AB} 
  & MAE        & 64.89 & \textbf{54.80} & 58.46 & \textbf{49.70} \\
  & PSNR       & 29.10 & \textbf{30.97} & 31.33 & \textbf{32.09} \\
  & MS-SSIM    & 0.91  & \textbf{0.91}           & 0.90  & \textbf{0.91} \\
\midrule
\multirow{3}{*}{HN} 
  & MAE        & \textbf{65.15} & 70.07          & 60.97 & \textbf{51.97} \\
  & PSNR       & \textbf{30.20} & 29.18 & 30.38 & \textbf{31.95} \\
  & MS-SSIM    & 0.94  & \textbf{0.95}  & 0.94  & \textbf{0.96} \\
\midrule
\multirow{3}{*}{TH} 
  & MAE        & 60.07 & \textbf{55.97} & 50.40 & \textbf{44.04} \\
  & PSNR       & 30.76 & \textbf{31.43} & \textbf{31.78} & 31.43 \\
  & MS-SSIM    & 0.94  & \textbf{0.95}  & 0.92  & \textbf{0.95} \\
\midrule
\multirow{3}{*}{Aggregated} 
  & MAE        & 63.37 & \textbf{60.28} & 56.61 & \textbf{48.57} \\
  & PSNR       & 30.02 & \textbf{30.53} & 31.16 & \textbf{31.82} \\
  & MS-SSIM    & 0.93  & \textbf{0.94}  & 0.92  & \textbf{0.94} \\
\bottomrule
\end{tabular}
\label{tab:local_split_task1_task2}
\end{table}

\begin{table}[H]
\centering
\caption{Comparison of synthesis performance between Baseline and IMPACT registration on the \textbf{Public} validation set.}
\begin{tabular}{lcc|cc}
\toprule
\multirow{2}{*}{\textbf{Metric}} & \multicolumn{2}{c|}{\textbf{Task 1}} & \multicolumn{2}{c}{\textbf{Task 2}} \\
 & \textbf{Baseline} & \textbf{IMPACT} & \textbf{Baseline} & \textbf{IMPACT} \\
\midrule
MAE  & \textbf{68.20} & 75.82 & \textbf{52.87} & 56.05 \\
PSNR & \textbf{29.81} & 28.70 & \textbf{32.36} & 31.65 \\
MS-SSIM  & \textbf{0.92} & 0.91 & \textbf{0.96} & 0.95 \\
Dice & \textbf{0.72} & 0.70 & \textbf{0.83} & 0.82 \\
HD95 & \textbf{8.42} & 8.89 & \textbf{5.40} & 5.41 \\
\bottomrule
\end{tabular}
\label{tab:public_split}
\end{table}

\begin{figure}[ht]
  \centering
  \includegraphics[width=0.6\textwidth]{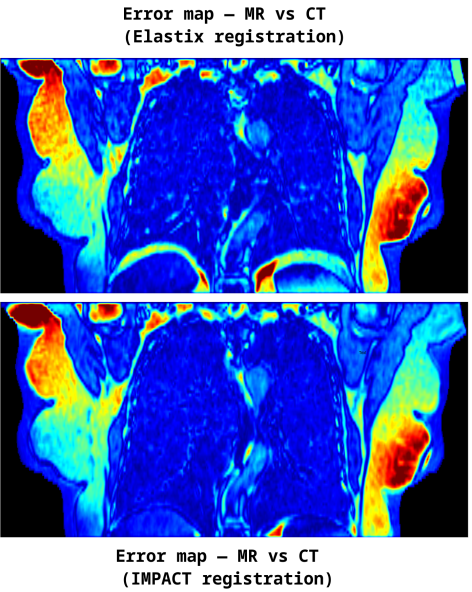}
\caption{Error maps for MR/CT registration using Elastix (top) and IMPACT (bottom).}
\label{fig:mr_ct_error_maps}
\end{figure}

\begin{figure}[ht!]
  \centering
  \includegraphics[width=0.8\textwidth]{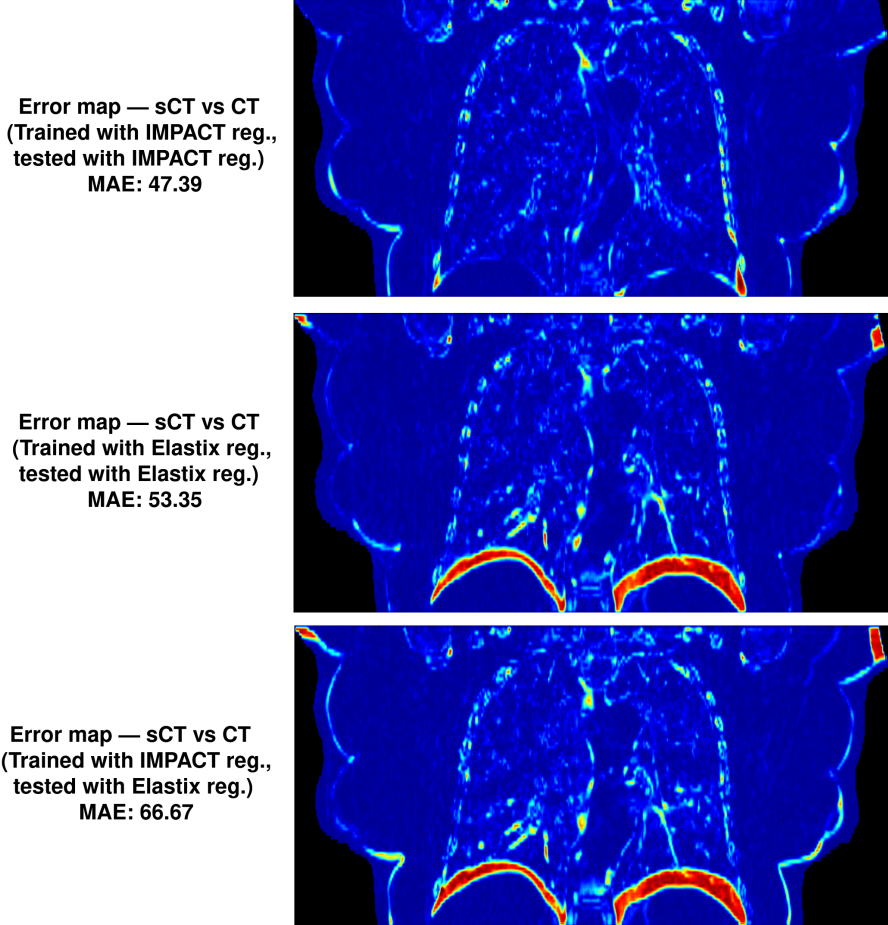}
\caption{Error maps of sCT vs CT for different training/test registration combinations.}
\label{fig:sct_ct_error_maps}
\end{figure}

\subsection{Results analysis.}
On the local test set, models trained with IMPACT-registered data outperformed those using baseline registration in both Task~1 and Task~2 (Table~\ref{tab:local_split_task1_task2}): MAE decreased, PSNR increased, and MS-SSIM improved or remained comparable.

Figure~\ref{fig:mr_ct_error_maps} shows that IMPACT achieves better MR/CT alignment. This leads to improved sCT synthesis (Figure~\ref{fig:sct_ct_error_maps}), with a MAE of 47.39 when training and testing both use IMPACT, compared to 53.35 with baseline registration.

In Task~1, gains were strongest in the Abdomen and Thorax. The Head and Neck showed slightly higher MAE with IMPACT but better MS-SSIM, indicating more consistent structural alignment. In Task~2, IMPACT outperformed the baseline across all regions and metrics, confirming its robustness.

Qualitative inspection also revealed that baseline-aligned data often introduced anatomical mismatches, especially in the lungs—leading models to “blur” structures and fill air spaces with tissue-like intensities to reduce pixel-wise error. This illustrates how poor registration can bias learning and compromise anatomy.

\subsection{Public validation set analysis.}
On the public validation set (Table~\ref{tab:public_split}), baseline-trained models outperformed those trained with IMPACT. This reflects a distributional bias from the validation pipeline, which favors models using the same registration. In supervised learning, where alignment defines the target, such bias distorts comparisons and does not benefit unsupervised approaches.

Thus, low error can reflect adaptation to registration artifacts rather than true anatomical mapping. This is evident when applying the IMPACT-trained model to Elastix-aligned inputs (MAE = 66.67 vs. 47.39; see Figure~\ref{fig:sct_ct_error_maps}), where performance collapses due to unseen distortions.

\section{Discussion} \label{sec:discussion}

Our study reveals a central yet often overlooked factor in supervised sCT generation: the quality of intermodal image registration. In supervised learning, voxel-level alignment between input and target modalities is implicitly assumed. Yet in real-world clinical scenarios, such alignment is rarely ensured due to inter-session variability, patient motion, and differences in acquisition protocols. As a result, multimodal images must be pre-aligned using registration algorithms, an essential but imperfect step that may introduce residual errors or spatial biases.

CNN-based encoder–decoder architectures excel at learning complex structural and intensity relationships between modalities. However, this very capacity makes them vulnerable to overfitting to registration artifacts. When exposed to misaligned training pairs, these models may exploit spatial inconsistencies to minimize image similarity losses, leading to deceptively strong metrics (MAE, PSNR, MS-SSIM) while compromising anatomical realism. This highlights the need to interpret such scores in light of registration fidelity, and to prioritize anatomically meaningful supervision.

IMPACT provides more accurate and anatomically consistent alignments than mutual information (MI) in multimodal registration tasks \cite{boussot2025impact}. In our experiments, these improved registrations appear to positively influence sCT synthesis, leading to higher image similarity metrics and more realistic anatomical structures. This supports the idea that registration quality plays an important role in supervised image synthesis.

Beyond registration, we introduce a perceptual loss (IMPACT-Synth) that uses a frozen segmentation backbone to preserve structural consistency during synthesis. Unlike traditional perceptual losses based on networks pre-trained on natural images for classification (e.g., VGG-based LPIPS), IMPACT-Synth relies on a domain-specific segmentation network. By combining pixel-wise objectives with segmentation features, it encourages anatomically plausible and structurally coherent outputs. Visual inspection reveals reduced blurring and sharper organ boundaries, underscoring the benefit of incorporating high-level semantic priors into the synthesis process.

IMPACT may therefore be a useful pre-alignment tool in pipelines where training performance depends on good spatial correspondence between modalities. Since the SynthRAD2025 dataset does not include registration files, we provide the IMPACT-based B-spline transformations as a resource to facilitate future developments on this dataset, publicly available at \url{https://huggingface.co/datasets/VBoussot/synthrad2025-impact-registration}.

Future work will focus on comparing supervised and unsupervised pipelines under fair conditions, limiting the influence of registration bias. In addition, we plan to systematically benchmark different pretrained segmentation models as perceptual feature extractors, against the VGG baseline.

\section{Author contributions} 
Conceptualization: V. Boussot, C. Hémon. Methodology: V. Boussot, C. Hémon. Software and experiments: V. Boussot. Supervision: J.-L. Dillenseger, J.-C. Nunes. Writing – original draft: V. Boussot, C. Hémon. Writing – review and editing: all authors. 

\section*{Acknowledgment}
The work presented in this article was supported by the Brittany Region through its Allocations de Recherche Doctorale framework and by the French National Research Agency as part of the VATSop project (ANR-20-CE19-0015). Additionally, it was supported by a PhD scholarship Grant from Elekta AB (C.Hémon). The authors have no relevant financial or non-financial interests to disclose. While preparing this work, the authors used ChatGPT to enhance the writing structure and refine grammar. After using these tools, the authors reviewed and edited the content as needed and took full responsibility for the publication’s content.

\bibliographystyle{splncs04}
\bibliography{references}

\end{document}